\documentclass{article}

\usepackage{arxiv}
\usepackage{float}  
\usepackage[utf8]{inputenc} 
\usepackage[T1]{fontenc}    
\usepackage{hyperref}       
\usepackage{url}            
\usepackage{booktabs}       
\usepackage{amsfonts}       
\usepackage{nicefrac}       
\usepackage{microtype}      
\usepackage{lipsum}
\usepackage[utf8]{inputenc}
\usepackage{graphicx}
\graphicspath{ {./images/} }
\usepackage{amsmath}

\DeclareMathOperator*{\argmin}{arg\,min}
\usepackage{subcaption}
\usepackage{tabto}
\usepackage[table,xcdraw]{xcolor} 

\title{Convolutional Kolmogorov-Arnold Networks}

\author{
  \parbox[t]{0.45\textwidth}{
    \centering
    \textbf{Alexander Dylan Bodner} \\
    \texttt{Universidad de San Andrés} \\
    \texttt{Buenos Aires, Argentina} \\
    \texttt{abodner@udesa.edu.ar}
  }
  \hfill
  \parbox[t]{0.45\textwidth}{
    \centering
    \textbf{Jack Natan Spolski} \\
    \texttt{Universidad de San Andrés} \\
    \texttt{Buenos Aires, Argentina} \\
    \texttt{jspolski@udesa.edu.ar}
  }
  \vspace{1cm}
  \\
  \parbox[t]{0.45\textwidth}{
    \centering
    \textbf{Antonio Santiago Tepsich} \\
    \texttt{Universidad de San Andrés} \\
    \texttt{Buenos Aires, Argentina} \\
    \texttt{atepsich@udesa.edu.ar}
  }
  \hfill
  \parbox[t]{0.45\textwidth}{
    \centering
    \textbf{Santiago Pourteau} \\
    \texttt{Universidad de San Andrés} \\
    \texttt{Buenos Aires, Argentina} \\
    \texttt{spourteau@udesa.edu.ar}
  }
}

\begin{document}
\maketitle
\begin{abstract}
In this paper, we present Convolutional Kolmogorov-Arnold Networks, a novel architecture that integrates the learnable spline-based activation functions of Kolmogorov-Arnold Networks (KANs) into convolutional layers. By replacing traditional fixed-weight kernels with learnable non-linear functions, Convolutional KANs offer a significant improvement in parameter efficiency and expressive power over standard Convolutional Neural Networks (CNNs). We empirically evaluate Convolutional KANs on the Fashion-MNIST dataset, demonstrating competitive accuracy with up to $50\%$ fewer parameters compared to baseline classic convolutions. This suggests that the KAN Convolution can effectively capture complex spatial relationships with fewer resources, offering a promising alternative for parameter-efficient deep learning models. 
\end{abstract}


\section{Introduction}

The field of deep learning is constantly changing, the fast improvement of architectures has helped the advancement of computer vision in tasks involving complex spatial data. Convolutional Neural Networks proposed by LeCun et al.\cite{LeCun1998} are widely used due to their ability to handle high dimensional data arrays such as images. Normally, these networks rely on linear transformations followed by an optional activation function in their convolutional layers to understand spatial relationships, which significantly reduced the number of parameters to capture complex patterns in images. 

Recent years have seen an increase in the integration of advanced mathematical theories into deep learning architectures which have helped neural networks in handling complex data structures. Kolmogorov-Arnold Networks (KANs) \cite{Ziming2024} are a promising alternative to Multi-Layer Perceptrons (MLPs)\cite{hornik1989} that use the Kolmogorov-Arnold theorem to integrate splines which is a key component of their architecture.
 
In light of these advancements, this paper explores the adaptation of KANs to convolutional layers, a common element in many CNN architectures used in computer vision. Traditional CNNs utilize fixed activation functions and linear transformations which, while effective, can benefit from the flexibility offered by KANs. By employing spline-based convolutional layers, as proposed in SplineCNN by M. Fey and J. E. Lenssen et al. \cite{Fey2018}, networks can capture non-linear relationships more effectively.

Throughout this paper, we begin with a high-level overview of the KAN architecture to set the stage for a comprehensive mathematical treatment of Convolutional KANs. We will provide a detailed examination of different Convolutional KANs architectures and benchmark their performance against traditional models, focusing on parameter efficiency within the  Fashion-MNIST dataset. Our hypothesis posits that Convolutional KANs, by leveraging spline-based layers, will require fewer parameters while achieving accuracy levels competitive with established benchmarks, potentially setting a new standard in neural network architectures for image-related tasks \cite{Denil2013}. For further exploration and practical application, the code for this layer and all the experiments is available at our GitHub repository: \href{https://github.com/AntonioTepsich/Convolutional-KANs}{GitHub/Convolutional-KANs}.

\section{Related work}

\textbf{Kolmogorov-Arnold theorem and neural networks}

The application of the Kolmogorov-Arnold theorem in neural networks marks a significant theoretical integration that enhances the expressiveness and efficiency of neural models. The theorem, which provides a way to represent any multivariate continuous function as a composition of univariate functions and additions, has been adapted in the design of Kolmogorov-Arnold Networks (KANs). KANs differ from traditional MLPs by replacing linear weight matrices with learnable splines, thus reducing the number of parameters required and potentially improving the generalization capabilities of the network \cite{Ziming2024}.

\textbf{Splines in Convolutional Neural Networks}

One noteworthy development in neural network architecture involves the use of splines, particularly in the context of Convolutional Neural Networks (CNNs). The SplineCNN, as proposed by M. Fey and J. E. Lenssen et al.\cite{Fey2018}, introduces spline-based convolutional layers that enhance the network's ability to capture non-linear relationships in the data. This approach is particularly effective in geometric deep learning, where the adaptability of splines plays a crucial role in handling non-Euclidean data.

A significant aspect of the method proposed by the authors is its treatment of images like those in the MNIST dataset, where it first processes the images by interpreting them as graphs before classification. This graph-based approach allows SplineCNN to handle irregular data structures effectively. Unlike SplineCNN, our Convolutional Kolmogorov-Arnold Networks (Convolutional KANs) apply spline functions directly on structured data such as images and matrices without needing their conversion into graphs.

\section{Kolmogorov-Arnold Networks (KANs)}
\label{sec:kan}
 In the evolving landscape of neural networks, the use of Kolmogorov-Arnold Networks (KANs) \cite{Ziming2024} presents an innovative approach to neural network design, based on the Kolmogorov-Arnold representation theorem \cite{Arnold1957}. This theorem states that any multivariate continuous function can be represented as a composition of univariate functions and addition operations. This foundational concept sets KANs apart from traditional Multi-Layer Perceptrons (MLPs) \cite{hornik1989}.

\subsection{Architecture}
 The core of KANs resides in their unique architecture. Unlike traditional MLPs that use fixed activation functions at nodes, KANs implement learnable activation functions on the network edges. This critical shift from static to dynamic node functions involves replacing conventional linear weight matrices with adaptive spline functions, which are parametrized and optimized during training. This allows for a more flexible and responsive model architecture that can dynamically adapt to complex data patterns.

 In more detail, the Kolmogorov-Arnold representation theorem posits that a continous multivariate function \( f(x_1, \ldots, x_n) \) can be expressed as:

\begin{equation*}
    f(x_1, \ldots, x_n) = \sum_{q=1}^{2n+1} \Phi_q \left( \sum_{p=1}^n \phi_{q,p}(x_p) \right)
\end{equation*}
Here, \( \phi_{q,p} \) are univariate functions mapping each input variable ($x_p$) such $\phi_{q,p} : [0,1] \to \mathbb{R}$, and  $\Phi_q$ : $\mathbb{R} \to \mathbb{R}$, univariate functions. 

KANs structure each layer as a matrix of these learnable 1D functions:
$$
\Phi = \{\phi_{q,p}\}, \quad p = 1, 2, \ldots, n_{in}, \quad q = 1, 2, \ldots, n_{out}
$$
Particularly, each function \( \phi_{q,p} \) can be defined as a B-spline, a type of spline function defined by a linear combination of basis splines, enhancing the network's ability to learn complex data representations. Here, \( n_{in} \) represents the number of input features to a particular layer, while \( n_{out} \) denotes the number of output features produced by that layer, reflecting the dimensionality transformations across the network layers. The activation functions \( \phi_{l,j,i} \) in this matrix are such learnable spline functions, expressed as:
\[
\text{spline}(x) = \sum_i c_i B_i(x), \quad c_i \text{ are trainable coefficients}
\]

This formulation allows each \( \phi_{l,j,i} \) to adapt its shape based on the data, offering unprecedented flexibility in how the network models interactions between inputs.

The overall structure of a KAN is analogous to stacking layers in MLPs, but with the enhancement of utilizing complex functional mappings instead of simple linear transformations and nonlinear activations:
$$
\text{KAN}(x) = (\Phi_{L-1} \circ \Phi_{L-2} \circ \cdots \circ \Phi_0)(x)
$$
Each layer's transformation, \( \Phi_l \), acts on the input \( x_l \) to produce the next layer's input \( x_{l+1} \), described as:
$$
x_{l+1} = \Phi_l(x_l) = \left( \begin{array}{ccc}
\phi_{l,1,1}(\cdot) & \cdots & \phi_{l,1,n_l}(\cdot) \\
\vdots & \ddots & \vdots \\
\phi_{l,n_{l+1},1}(\cdot) & \cdots & \phi_{l,n_{l+1},n_l}(\cdot)
\end{array} \right) x_l
$$
where each activation function \( \phi_{l,j,i} \) is a spline, providing a rich, adaptable response surface to model inputs.

\subsection{Motivation for Using KANs}

The motivation for using the architecture of KANs, with learnable activation functions on edges, enhances their expressive power and efficiency. By replacing linear weight matrices with spline functions, KANs reduce the number of parameters needed to achieve high accuracy, leading to faster convergence and better generalization.

\section{Convolutional Kolmogorov-Arnold Networks}
Convolutional Kolmogorov-Arnold Networks are similar to CNNs. The difference is that the Convolutional Layers are replaced by KAN Convolutional Layers and after flattening, one can either have a KAN or a MLP. The main strength of the Convolutional KANs is its requirement for significantly fewer parameters compared to other architectures. This is given by the construction of this networks, because B-Splines are able to smoothly represent aribtrary activation functions that will not be found using a ReLU in between convolutions. 

\subsection{KAN Convolutions}
In computer vision, convolutions are a foundational operation in Convolutional Neural Networks (CNNs), where a kernel or filter slides across the input, computing dot products at each position to extract spatial features. Unlike traditional CNNs, where kernels consist of fixed weights, KAN Convolutions use kernels composed of learnable non-linear functions, denoted by $\phi$, which are parameterized using B-splines. This approach allows each kernel element to adapt dynamically during training, enabling greater flexibility and expressiveness. Formally, each element of the kernel is defined as:
\newline
\begin{equation}\label{eqn:b_spline}
    \phi = w_1 \cdot spline(x) + w_2 \cdot silu(x) \quad 
\end{equation}
 
In a KAN Convolution, the kernel slides over the image and applies the corresponding activation function, $\phi_{ij}$ to the corresponding pixel, $a_{kl}$ and calculates the output pixel as the sum of $\phi_{ij}(a_{kl})$. Let K be a KAN kernel $\in \mathbb{R}^{N\times M}$, and an image as a matrix as:

    \begin{equation}
        \text{Image} = 
        \begin{bmatrix}
            a_{11} & a_{12} & \cdots & a_{1p} \\
            a_{21} & a_{22} & \cdots & a_{2p} \\
            \vdots & \vdots & \ddots & \vdots \\
            a_{m1} & a_{m2} & \cdots & a_{mp}
        \end{bmatrix}
    \end{equation}

Then a KAN Convolutions is defined as follows in the Equation \ref{eqn:kan_convolution_eq}
\begin{equation}
 (\text{Image} \ast K)_{i,j} = \sum_{k=1}^{N}\sum_{l=1}^{M} \phi_{kl}(a_{i+k,j+l})
  \label{eqn:kan_convolution_eq}
\end{equation}
Lets see an example in matrix form:

\[
    \text{KAN Kernel} = 
    \begin{bmatrix}
    \phi_{11} & \phi_{12} \\
    \phi_{21} & \phi_{22}
    \end{bmatrix}
\]

\begin{equation}
\text{Image} \ast \text{KAN Kernel} =
\begin{bmatrix}
    \phi_{11}( a_{11}) + \phi_{12} (a_{12}) + \phi_{21} (a_{21}) + \phi_{22} (a_{22}) & \cdots & r_{1(p-1)} \\
    \phi_{11} (a_{21}) + \phi_{12}  (a_{22}) + \phi_{21} (a_{31}) + \phi_{22}  (a_{32}) & \cdots & r_{2(p-1)} \\
    \vdots & \ddots & \vdots \\
    \phi_{11} (a_{m1}) + \phi_{12}  (a_{m2}) + \phi_{21} (a_{(m+1)1}) + \phi_{22}  (a_{(m+1)2}) & \cdots & r_{m(p-1)}
\end{bmatrix}
\end{equation}

\subsection{KAN filters theoretical advantages}

The early application of different nonlinearities to each pixel makes KAN kernels able to do things that classic kernels cannot. Because each pixel can be transformed nonlinearly before being summed, the resulting filter can encode complex, context-dependent behaviors in a single pass. By contrast, classic linear kernels merely combine pixel intensities through fixed, constant weights. While multiple linear filters can be stacked or summed, they still cannot implement conditional or piecewise logic at the pixel level without separate processing steps. KAN filters, however, can carry out thresholding, adaptive smoothing, and edge detection all at once. 

Let's take a look at some of these examples to illustrate the unique capabilities of KAN filters:
\subsubsection{Threshold + Convolution}
\label{sec:threshold_linear}

Suppose we want to:
\begin{enumerate}

    \item \textbf{Apply a threshold} on specific pixels (e.g., the center pixel).
    \item \textbf{Linearly weight} each pixel's intensity. This would also apply to other desired operations like Sobel or Laplacian filters.
\end{enumerate}
A single classical filter cannot perform a threshold in the same pass (that requires a global post-sum activation or a separate step). In KAN form, however, we define each $\phi_{i,j}$ as:
\[
\phi_{i,j}(x) \;=\; 
\begin{cases}
w_{i,j} \, x, & \text{if $(i,j)$ is not the center},\\[6pt]
\max(0,x-1) & \text{if $(i,j)$ is center}.
\end{cases}
\]
Here, the second function looks like a displaced ReLU, which doesn't let any  positive value between 0 and 1 pass, and then progressively activates. To clarify, this would not be exactly the learned function because we can only learn smooth functions, but it can be sufficiently close in practice.

Summing $\phi_{i,j}(x_{i,j})$ over all $i,j$ yields a final image where we both apply a linear weighting \emph{and} smooth thresholding to the center pixel in one pass.

\subsubsection{ Piecewise Brightness-Based Boost}
\label{sec:piecewise_boost}

Although $\phi_{i,j}$ cannot see neighboring pixels, it can still apply a different multiplier depending on whether its own intensity is above or below a threshold. For instance:
\[
\phi_{i,j}(x)
\;=\;
\begin{cases}
\alpha_{\text{bright}} \,\cdot\, x, & \text{if } x \ge T, \\[6pt]
\alpha_{\text{dark}} \,\cdot\, x,   & \text{otherwise}.
\end{cases}
\]
Here:
\begin{itemize}
    \item $\alpha_{\text{bright}}$ and $\alpha_{\text{dark}}$ are different gains for bright vs.\ dark pixel intensities.
    \item If $x_{i,j} \ge T$, we ``boost'' it more ($\alpha_{\text{bright}}$), else we ``boost'' less ($\alpha_{\text{dark}}$).
\end{itemize}
Summing $\phi_{i,j}(x_{i,j})$ over $i,j$ yields an output that \emph{conditionally scales} each pixel based on its own brightness---something no single linear filter can replicate. For instance, if $\alpha_{\text{bright}} > \alpha_{\text{dark}}$, bright regions get emphasized more strongly than dark regions in \emph{one} pass.

\subsubsection{Key takeaways}

KAN filters, even with only per-pixel access, allow us to mix linear and nonlinear operations (thresholding, piecewise gains, small saturations, etc.) \emph{all in a single pass}—whereas a single classic filter must remain purely linear, and any nonlinear steps would typically happen \emph{after} that pass and without the possibility of letting gradient descent learn the activation function. Also, if we would want to achieve the same results as the KAN Convolution, we would need to apply pixel operations before the convolution, which is not typically done in CNNs.

\subsection{Grid extension and update}
The grid refers to the set of points which we want to discretize. It is initialized with a previously defined interval and control points, but during training, some of the input variables to each $\phi$ might get out of range from the grid limits. To tackle this we extended the grid to be able to capture the variables which escape the original limits. The method to extend the grid described in the original KAN paper and consists in the following optimization problem.

\begin{equation}
\{c_j' \} = \argmin_{c'_j} \mathbb{E}_{x\sim p(x)}[ \sum _{j=0}^{G_2+k-1}c_{j}' B(x')_{j}' - \sum _{j=0}^{G_1+k-1}c_{j} B(x)_{j}]
\label{eqn:grid_ext}
\end{equation}
Where $G_1$ is the previous grid size, $G_2$ is the new grid size and k is the B Spline degree.

\begin{figure}[h]
     \centering
     \includegraphics[width=1\linewidth]{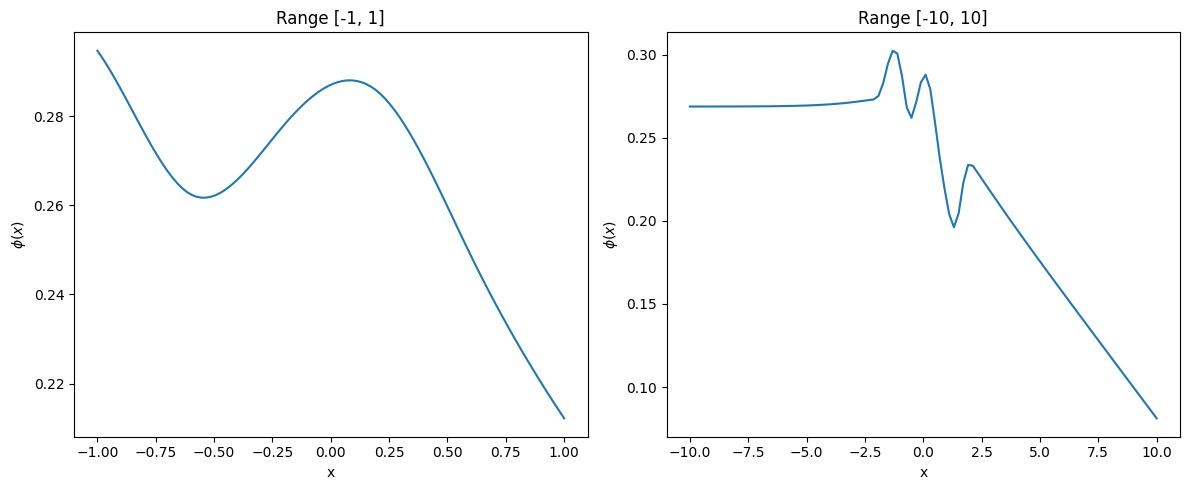}
     \caption{Splines learned by the first convolution at the first position for different ranges. The left plot shows the spline learned within the range \([-1, 1]\), while the right plot shows the spline learned within the range \([-10, 10]\). The SILU (Sigmoid Linear Unit) function is added to the spline across the entire range, but the spline is only defined within \([-1, 1]\). Thus, outside this range, the SILU function predominates.}
     \label{fig:range_vis}
 \end{figure}

While testing the models we effectively verified that the output variables of a KAN Convolutional Layer weren't bounded to the default grid range of $[-1,1]$. This is a problem, especially when using multiple convolutional layers since the input of a convolutional layer should be in the range that the B-Spline operates in so that the "learning" is done by the splines and not the weight that modifies the SiLu. To solve this issue, during training each time an input falls outside the grid range, the grid is updated. This consists of maintaining spline shape between the original grid size and maintaining the same amount of control points, and extending the spline to a range that contains the input. Another solution for this issue is the implementation of batch normalization, after each convolutional layer a batch normalization layer is applied. This approach adds very few learnable  $\mu$ and $\sigma$ parameters that Standardize  the inputs to the layer to $\mu = 0$ and $\sigma = 1$. This ensures that most, but not all, of the outputs are between in range. As is seen in Figure \ref{fig:range_vis}, when the input is out of the Spline range, the layer will act as as SiLu activation as is expected, so if the range is not updated and most of inputs land out of range, a KAN will not differ from an MLP that uses SiLu activations.

 Upon analyzing the splines learned by the network in different convolutions, we did not find any recognizable pattern. The behavior of the splines varies significantly across different convolutional layers and positions, indicating that the learning process is highly context-dependent and does not conform to a simple, uniform structure.
\subsection{Parameters in a Convolutional KAN}
As previously mentioned, the amount of parameters is one of the main advantages of using Convolutional KANs. With $\phi$ defined as in Equation \ref{eqn:b_spline}, the parameters for each $\phi$ are the two weights, $w_1$ and $w_2$, together with the control points which can be adjusted to change the shape of each spline. Therefore there are $gridsize + 2$ parameters for each $\phi$. Let the convolution kernel be of size $K\times K$, in total we have K²(gridsize + 2) parameters for each Convolutional KAN layer, compared to only K² for a CNN convolutional layer. In our experiments the gridsize is typically between k and k², with k tending to be a small value between 2 and 16. 

In the convolutional layers, Convolutional KANs have more parameters, but as they utilize splines, they have more adaptability to process the spatial information and thus require less amount of fully connected layers which significantly increase the amount of parameters. That is where the advantage of utilizing splines really shows, we are able to reduce the amount of non-convolutional layers and thus reduce the parameter count.

\section{Experiments}
In this section we explain the different experiments we conducted to analyze the performance of different models that use KAN Convolutional Layers against a classical convolutional neural-network. 

During experimentation we used two datasets, MNIST and Fashion MNIST. Doing all the experimentation in both datasets would double the time and cost of the experiments, difficulting to run all the proper experiments of different architectures. As Fashion MNIST presents some more complexity than MNIST, we decided to tune hiperparameters and present results only on Fashion MNIST. 

We proposed architectures that use a mix between Fully Connected (MLP), KAN, KAN Convolutional and Convolutional layers. The following Figure \ref{fig:CKAN-Architectures} shows the different architectures used that have KANs:
  \begin{figure}[H]
     \centering 
     \includegraphics[width=0.9 \linewidth]{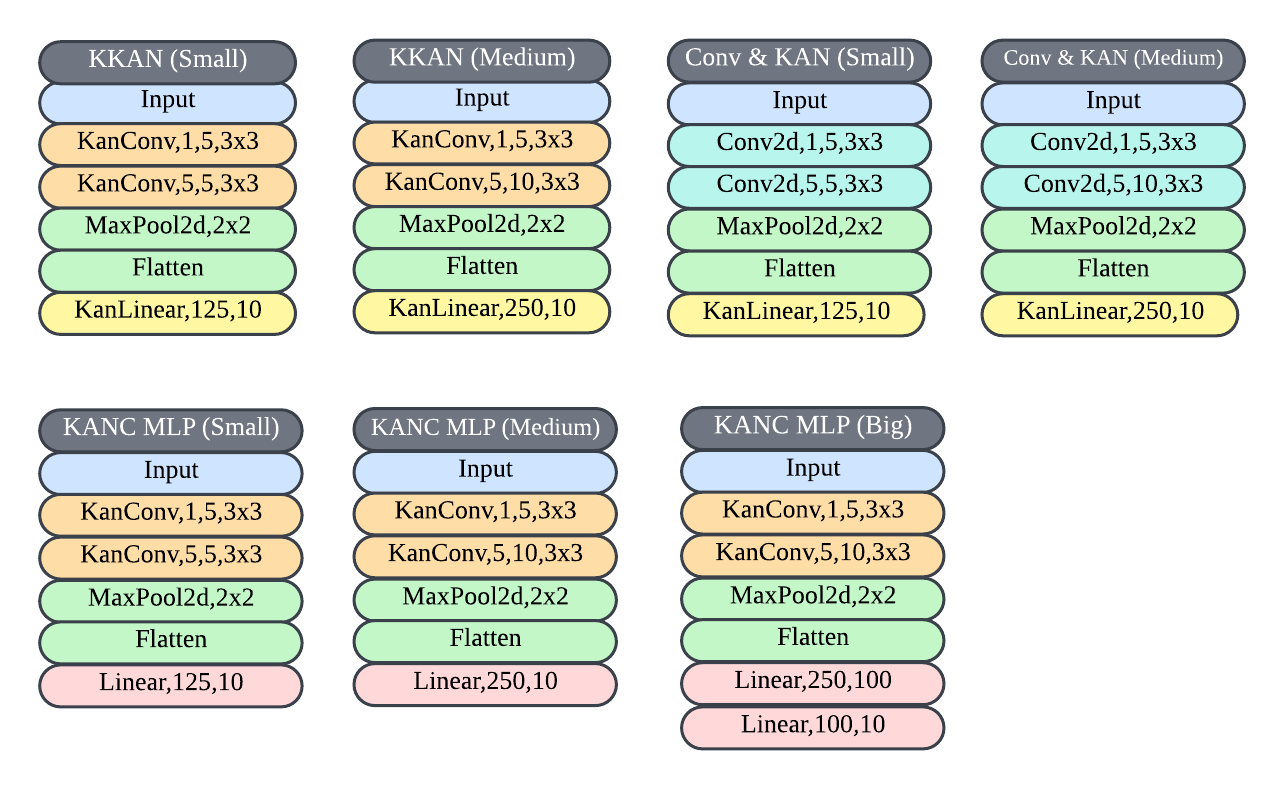}
     \caption{ KAN Architectures used in experiments. The Max Pooling layers are done after every Convolutional Layer, but for simplicity sake of the scheme we decided to show it only at the end. Every architecture has at the end a Log Softmax layer.}
     \label{fig:CKAN-Architectures}
 \end{figure}
 
Figure \ref{fig:Standard-Architectures} shows the different Standard architectures used in the experiments:
  \begin{figure}[H]
     \centering 
     \includegraphics[width=0.7 \linewidth]{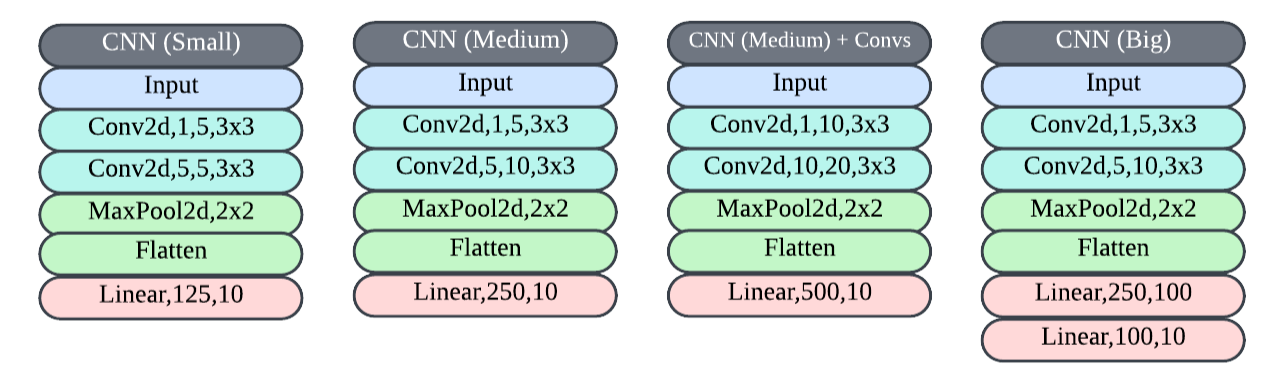}
     \caption{Standard Architectures used in experiments. The Max Pooling layers are done after every Convolutional Layer, but for simplicity sake of the scheme we decided to show it only at the end. Every architecture has at the end a Log Softmax layer.}
     \label{fig:Standard-Architectures}
 \end{figure}

We conducted a hyperparameter search using grid search, testing 8 different combinations for each model. For KAN models, we performed different grid searches for KAN grid sizes of 10 and 20 (the grid size for the B-Spline), as parameter count increases significantly with grid size and we wanted to compare how the accuracy changes when grid size changes. Another possible hyperparameter is the B-Spline degree. We decided to not tune this hyperparameter and chose it as degree 3, as KAN authors \cite{Ziming2024} suggest as default and because of the wide usage of cubic splines. It is important to define the degree sufficiently small to not overfit.

Because of the long training times of these models, we opted not to do a K-Fold Cross Validation, but to split the training set into train and valid sets and tune hyperparameters with those sets. Once we find the best hyperparameter sets with the previous datasets, we join train and valid together, train a model with train and valid sets with the optimal hyperparameters and finally report the results using test set.

The possible values for the hiperparamters are defined in Table \ref{tab:hyperparameters}
\begin{table}[h!]
    \centering
    \caption{Possible Values for Each Hyperparameter}
    \begin{tabular}{ll}
        \toprule
        \textbf{Hyperparameter} & \textbf{Values} \\
        \midrule
        Learning Rate      & \{ 0.0001, 0.0005, 0.001\} \\
        Weight Decay & \{0, 0.00001, 0.0001\} \\
        Batch Size & \{32, 64, 128\} \\
        \bottomrule
    \end{tabular}
    \label{tab:hyperparameters}
\end{table}

In addition to these possible values, the epochs are also a learned hiperparameter.

\subsection{Loss}

For every model that we trained, the Categorical Cross Entropy loss was used as base,  but for KAN models, there are 2 additional regularization terms proposed in the KAN paper \cite{Ziming2024}.

Cross entropy loss is defined as:

\begin{equation}
L_{ce}= -\sum_{i=1}^N \sum_{c=1}^Cy_{i,c}\log(p_{i,c})
\label{eqn:cross_entropy}
\end{equation}

KAN loss with regularization is defined as:

\begin{equation}
L_{reg}= L_{ce} + \lambda(\mu_1\sum_{l=1}^{L-1}|\Phi_l|_1 +\mu_2\sum_{l=1}^{L-1} S(\Phi_l))
\label{eqn:loss_regularized}
\end{equation}

With:

\begin{equation}
 S(\Phi_l)) =  -\sum_{i=1}^{n_{in}}\sum_{j=1}^{n_{out}}\frac{|\phi_{i,j}|}{|\Phi|_1}log(\frac{|\phi_{i,j}|}{|\Phi|_1})
\label{eqn:S}
\end{equation}

In our early experiments we found out that the trained models performed better and trained faster without the regularization terms, so in our experiments we ended defining $\lambda = 0$.

\subsection{Hardware}
For conducting the hyperparameter tuning for all the models, we used a 'g2-standard-4' Google Cloud instance, which has 4 VCPUS, 2 cores, 16 RAM and a NVIDIA L4 GPU.  

 \section{Results}
This section displays an analysis of the performance of the different proposed models in the previously described experiments. To obtain the hyperparameters and final models, the training time was of almost 2 and a half days. Table \ref{tab:model_performances_test} presents a comparison of accuracy, precision, recall, F1 score, parameter count, and training time per epoch for the proposed models tested on the Fashion MNIST Dataset.

\begin{figure}[H]
     \centering
     \includegraphics[width=1\linewidth]{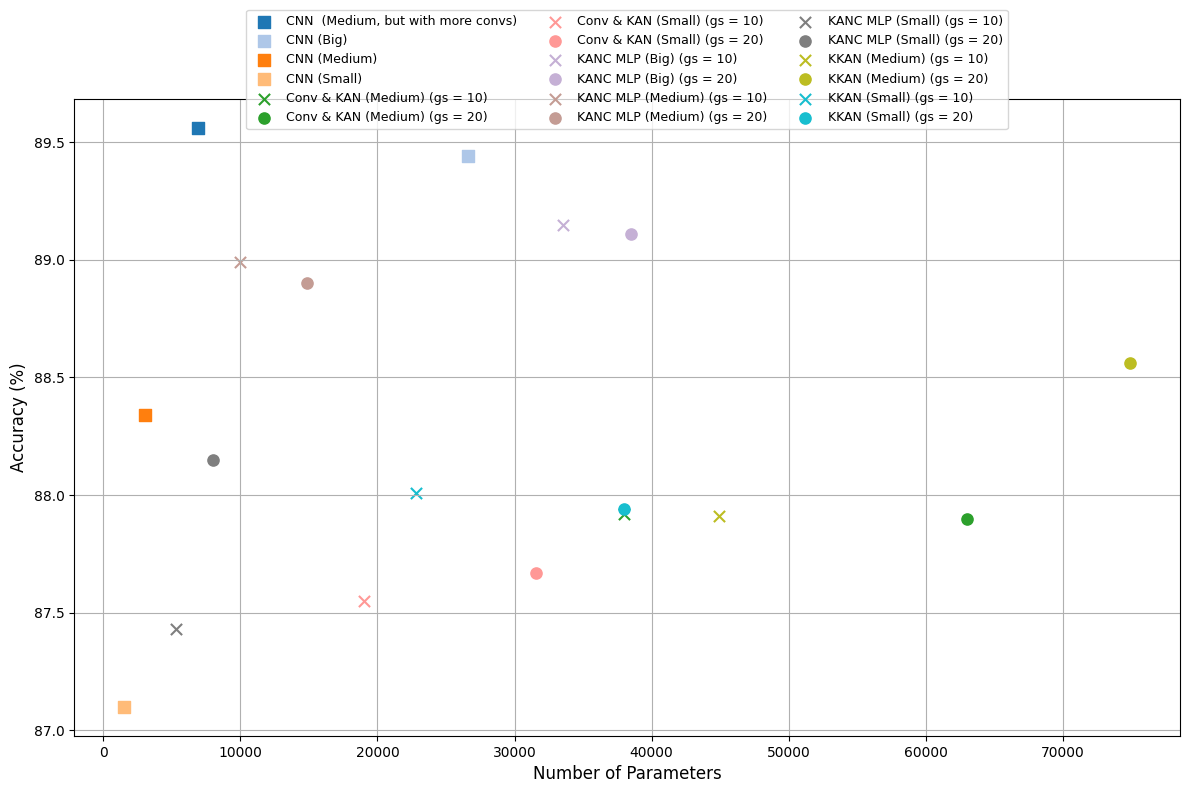}
     \caption{Parameter count vs Accuracy in Fashion-MNIST dataset.}
     \label{fig:fashion_mnist_plot}
 \end{figure}

\begin{table}[H]
\centering
\begin{tabular}{lcccccc}
\toprule
\textbf{Model} & \textbf{Accuracy} & \textbf{Precision} & \textbf{Recall} & \textbf{F1 Score} & \textbf{\#Params} & \textbf{Minutes per epoch} \\
\midrule
CNN (Medium, but with more convs) & 89.56\% & 89.60\% & 89.56\% & 89.55\% & 6.93K & 0.2103 \\
CNN (Big) & 89.44\% & 89.40\% & 89.44\% & 89.39\% & 26.62K & 0.2114 \\
CNN (Medium) & 88.34\% & 88.20\% & 88.34\% & 88.22\% & 3.02K & 0.2103 \\
CNN (Small) & 87.10\% & 87.08\% & 87.10\% & 87.01\% & 1.54K & 0.2328 \\
Conv \& KAN (Medium) (gs = 10) & 87.92\% & 87.84\% & 87.92\% & 87.80\% & 38.01K & 0.2306 \\
Conv \& KAN (Medium) (gs = 20) & 87.90\% & 88.08\% & 87.90\% & 87.95\% & 63.01K & 0.2266 \\
Conv \& KAN (Small) (gs = 10) & 87.55\% & 87.37\% & 87.55\% & 87.42\% & 19.03K & 0.2303 \\
Conv \& KAN (Small) (gs = 20) & 87.67\% & 87.74\% & 87.67\% & 87.59\% & 31.53K & 0.2254 \\
KANC MLP (Big) (gs = 10) & 89.15\% & 89.22\% & 89.15\% & 89.14\% & 33.54K & 1.6523 \\
KANC MLP (Big) (gs = 20) & 89.11\% & 89.05\% & 89.11\% & 89.06\% & 38.48K & 2.4633 \\
KANC MLP (Medium) (gs = 10) & 88.99\% & 88.97\% & 88.99\% & 88.96\% & 9.94K & 1.6441 \\
KANC MLP (Medium) (gs = 20) & 88.90\% & 88.85\% & 88.90\% & 88.83\% & 14.89K & 2.4615 \\
KANC MLP (Small) (gs = 10) & 87.43\% & 87.41\% & 87.43\% & 87.41\% & 5.31K & 1.1704 \\
KANC MLP (Small) (gs = 20) & 88.15\% & 88.02\% & 88.15\% & 87.94\% & 8.01K & 1.6262 \\
KKAN (Medium) (gs = 10) & 87.91\% & 88.37\% & 87.91\% & 87.99\% & 44.93K & 1.6425 \\
KKAN (Medium) (gs = 20) & 88.56\% & 88.52\% & 88.56\% & 88.52\% & 74.88K & 2.4753 \\
KKAN (Small) (gs = 10) & 88.01\% & 87.87\% & 88.01\% & 87.76\% & 22.80K & 1.1599 \\
KKAN (Small) (gs = 20) & 87.94\% & 87.80\% & 87.94\% & 87.72\% & 38.00K & 1.6336 \\
\bottomrule
\end{tabular}
\caption{Comparison of test accuracy, precision, recall, F1 score, parameter count, and training time per epoch for various models. KKAN and KANC MLP architectures show competitive performance, balancing accuracy and parameter efficiency.}
\label{tab:model_performances_test}
\end{table}
Results of Table \ref{tab:model_performances_test} , can also be visualized in Figure \ref{fig:fashion_mnist_plot} which illustrates the relationship between parameter count and accuracy achieved by these models, providing a graphical representation of the results.

Following the results analysis, Figure \ref{fig:fashion_mnist_plot} and Table \ref{tab:model_performances_test} illustrate the relationship between parameter counts versus accuracy for the described models but applied to the Fashion MNIST dataset. In the experiments with this dataset, we find that: 

In the smaller models, we find that both using MLPs or KANs after the flatten, KAN convolutions outperform  classic convolutions. This are the Small and Medium models and here the fair comparisons are:
\begin{itemize}

\item KANC MLP (Small) vs CNN(Small).

\item KANC MLP (Medium) vs CNN(Medium).
\item Normal Convs \& KKAN (Small) vs KKAN (Small).
\item Normal Convs \& KKAN (Medium) vs KKAN (Medium).
\end{itemize}

Although KANC MLP and CNN of equal sizes have the same architecture and only changes the type of convolution, KAN convolutions provide a considerable amount of parameters. Ignoring the difference of parameters, as said before, within the same architecture, both Small and Medium KANC MLPs outperform CNNs. But, for example, KANC MLP Small has more trainable parameters than CNN Medium and had slightly less accuracy ($88.15\%$ vs $88.34\%$). While it has more parameters, it has less convolution kernels, so in this case, it seems that the added expressibility of KAN Convolutions does not compensate the number of different filters. 

However, when we increase the depth of the MLPs, classic convolutions show a small advantage over KAN Convolutions. This can be seen in the case of CNN (Big) vs KANC MLP (Big), where CNN has $89.44\%$ and KANC MLP $89.15\%$. This can be explained by the fact that the learning could be done by the MLP, and maybe KAN Convolutions are compressing more the information, losing data that the MLP might need. Also, we tried a model with even more convolutions, which is CNN(Medium, but with more convs), and shows even better accuracy than CNN (Big), which has less convolutions, but one more fully connected layer and because of that, more parameters. 

In the case where we use a KAN Network after the convolutions, using KAN convolutions (KKAN Small) gets $87.67\%$ accuracy, while using classic convolutions (Conv \& KAN Small) gets $88.01\%$. Conv \& KAN is between 6 and 7 times faster to train (depending on the grid size), but KKAN Small achieves slightly better accuracy than Conv \& KAN Medium ($87.92\%$) and with almost half parameters (when Conv KAN Medium has grid size 20 and KKAN grid size 10). Needless to say, KKAN Medium, which only increases the amount of convolutions, surpasses this accuracy with grid size 20, which achieves $88.56\%$, but parameter count increases from 38000 to 74875 (Because of the increased number of convolutions and KAN Neurons).

In the current experiments, adding KAN kernels keeping the same number of Convolutional layers seem to faster reach a limit on the accuracy increase, while with classic convolutions it seems to be necessary to achieve a higher accuracy. One explanation to this might be the fact that KAN Kernels are more expressible and one kernel could be learning what it might take several classic kernels.

In addition, we can verify that grid size can change abruptly the model accuracy. For achieving the best performance this parameter should be tuned for every KAN layer, although this process demands significantly greater computational resources. In the cases of KKAN Small, KANC MLP Medium and Big and both Conv \& KANs, no accuracy was gained by increasing grid size from 10 to 20, and in some instances, accuracy was even reduced. However, in models such as KANC MLP Small and KKAN Medium, accuracy improved by over $0.5\%$, which represents a noticeable gain given the minimal accuracy differences across these models.

\section{Conclusions}
In this paper we proposed a new way to adapt the idea of learning splines proposed in Kolmogorov-Arnold Networks to convolutional layers used widely in Computer Vision. We implemented a KAN Convolutional Layer that uses a kernel made of learnable non-linear functions using B-Splines. We have found out that with equal architectures, KAN Convolutions seem to ``learn more", showing better accuracy in the analogous models except in the ``Big" ones, where the difference was that we added more fully connected layers, which may be doing the learning instead of the convolutions. 

When using KANs after the flatten, KAN Convolutions achieve better accuracy than classic convolutions, even using half the parameters. This is seen in the comparisons between KKAN and Normal Conv \& KAN. 

When trying a MLP, KAN convolutions achieve higher accuracy in the small models, but when having a 2 layer MLPs, the classic CNNs win by $0.41\%$ with $\sim 26.62$k parameters. While KAN Convolutions seem to learn more per kernel, we have to consider that each KAN kernel has much more parameters. So comparing the same architectures gives an advantage in terms of expressibility to KAN Convolutions. 

A key factor found is that KANs seem to maintain accuracy with lower parameter count, KANC MLP (Medium) achieves $88.99\%$ with approximately $9k$ parameters, in comparison to CNN (Big), which reaches $89.44\%$ with $26k$ parameters. But the training time is almost 8 times slower with the current implementations of KANs and its derivatives.  

Additionally, MLPs show a better performance than KANs for image classification tasks. In the ``Small" and ``Medium" cases, when using KAN Convolutions, using MLPs after the flatten gives both better accuracy and smaller parameter counts than using KANs. As expected, KKANs achieve slightly better accuracy than its equal CNNs, but the parameter count difference is too high (22k vs 1.5k and 38k vs 3k), without a corresponding improvement in accuracy ($0.91\%$ increase in accuracy for the Small case and $0.22\%$ for the Medium case). Since the parameters of KAN layers grow much faster than those of MLPs, with no significant gain in accuracy, KANs might not be suitable as dense layers, especially because CNNs require a large number of neurons after the flattening layer.

Based on the original KAN paper \cite{Ziming2024} proposal that KANs are more interpretable, we have tried to find a way to interpret this new type of convolutions, but at the moment, we have not found any clear way to visualize the B-Splines learned in each pixel of the convolutions besides that KAN Convolutions are able by constructions to learn more filters in one kernel. Given that we are working on images, it seems that the classic approach of visualizing what the filter does to the image seems the most 'human' way to get a sense of what is being learned.

The investigation leads us to understand the limitations of this new and promising idea. These limitations are similar to those presented in the original KAN paper. The KAN Linear Layer and our KAN Convolutional Layer are new and need to be optimized before they are able to be scaled properly, as shown in the time per epoch metrics. This paper is a starting point for integrating KANs into computer vision, and shows that Convolutional KANs have the potential to be an alternative to Convolutional Neural Networks. 
\section{Limitations and future work }

The current implementation of KANs with B-Splines is considerably slow due to its impossibility of being GPU parallelizable, making it very difficult to apply KANs in real world problems. Many authors are working on solutions to this by replacing B-Splines by other function approximators, such as Radial Basis Function \cite{radialbasis}, which are GPU parallelizable and open up a wide range of possibilities with KANs. 

Additionally, similar experiments must be conducted in more complex datasets such as CIFAR-10 or ImageNet to verify if the results found in this paper hold on scale.

Finally, interpretability is one of the strong points of KANs as mentioned in the original KAN paper. There is still place to work on that area, evaluating if there is a way to interpret KAN convolutions and if there are efficacious ways of pruning KAN convolutions.

\nocite{*}
\bibliographystyle{plain}  

\bibliography{references}	

\end{document}